\documentclass[twocolumn]{cinc}

\usepackage{graphicx, xurl, hyperref}
\usepackage{graphicx}
\usepackage{algorithm}
\usepackage{algorithmic}
\usepackage{amsmath,amssymb}
\usepackage{bm}
\usepackage{comment}
\usepackage[T1]{fontenc}

\title{Reduced-Lead ECG Classifier Model \\ Trained with DivideMix and Model Ensemble}
\author{Hiroshi Seki\textsuperscript{1}, Takashi Nakano\textsuperscript{1}, Koshiro Ikeda\textsuperscript{1}, Shinji Hirooka\textsuperscript{1}, Takaaki Kawasaki\textsuperscript{1}, \\  Mitsutomo Yamada\textsuperscript{1}, Shumpei Saito\textsuperscript{1}, Toshitaka Yamakawa\textsuperscript{2}, and Shimpei Ogawa\textsuperscript{1} \\ \ \\
\textsuperscript{1}AMI inc., Kagoshima, Japan 
\textsuperscript{2}Kumamoto University, Kumamoto, Japan}

\begin{document}
\maketitle

\begin{abstract}

Automatic diagnosis of multiple cardiac abnormalities from reduced-lead electrocardiogram (ECG) data is challenging.
One of the reasons for this is the difficulty of defining labels from standard 12-lead data.
Reduced-lead ECG data usually do not have identical characteristics of cardiac abnormalities because of the noisy label problem.
Thus, there is an inconsistency in the annotated labels between the reduced-lead and 12-lead ECG data.
To solve this, we propose deep neural network (DNN)-based ECG classifier models that incorporate DivideMix and stochastic weight averaging (SWA).
DivideMix was used to refine the noisy label by using two separate models. 
Besides DivideMix, we used a model ensemble technique, SWA, which also focuses on the noisy label problem, to enhance the effect of the models generated by DivideMix. 
Our classifiers (ami\_kagoshima) received scores of 0.49, 0.47, 0.48, 0.47, and 0.47 (ranked 9th, 10th, 10th, 11th, and 10th, respectively, out of 39 teams) for the 12-lead, 6-lead, 4-lead, 3-lead, and 2-lead versions, respectively, of the hidden test set with the challenge evaluation metric.
We obtained the scores of 0.701, 0.686, 0.693, 0.693, and 0.685 on the 10-fold cross validation, and 0.623, 0.593, 0.606, 0.612, and 0.601 on the hidden validation set for each lead combination.

\end{abstract}


\section{Introduction}

Cardiovascular disease is a leading cause of global mortality~\cite{Virani2021}.
As the electrocardiogram (ECG) can record the electrical activity of the heart non-invasively, there are a lot of studies on the automatic diagnosis of cardiac abnormalities from ECG.
PhysioNet/Computing in Cardiology Challenge 2021 focuses on the classification of cardiac abnormalities from reduced-lead ECGs~\cite{2020ChallengePMEA, 2021ChallengeCinC}. 

Real-world data are annotated by multiple human labelers with different skill levels. 
The annotation quality harms the performance of machine learning.
The annotation quality is also affected by different annotation rules of each hospital.
Therefore, there are many works on the noisy-label problem to train a robust model from noisy real-world datasets~\cite{DBLP:journals/corr/abs-1912-02911, li2020dividemix, izmailov2018averaging}.

Likewise, the reduced-lead ECG classification can be regarded as the noisy label problem because the reduction of certain ECG leads hinders the detection of important characteristics of cardiac abnormalities.

In this paper, we propose DNN-based ECG classifier models that are robust to annotation inconsistency.
These models incorporate DivideMix~\cite{li2020dividemix} and stochastic weight averaging (SWA)~\cite{izmailov2018averaging}.
We used {\sl EfficientNet}~\cite{tan2021efficientnetv2} which consists of 1D-CNN (convolutional neural network) for multi-label classification.

\section{{\hspace{-6mm}}Multi-class Classification with DivideMix}
\subsection{Base Classifier}
A multi-class classifier based on the DNN takes $|\mathcal{C}|$-dimensional ECG time-series data as input $X_c=(x_{c, t} \in {\mathbb{R}}^{|\mathcal{C}|}|t=1, \dots, T)$, and predicts the probabilities $Y=(y_1, \cdots, y_{N})$ 
where $T$ is the sequence length, $N$ is the number of diagnoses, and $y_i$ is the $i$-th label that takes 0 (negative) or 1 (positive).
The dimension $\mathcal{C}$ represents the lead combination.
In this study, we trained the classifier models to identify diagnoses from reduced-lead ECG sets: $c\in (\mathcal{C}_{\text 2}, \mathcal{C}_{\text 3}, \mathcal{C}_{\text 4}, \mathcal{C}_{\text 6}, \mathcal{C}_{\text{12}})$, where
$\mathcal{C}_{\text 2}=(\text{I, II})$, 
$\mathcal{C}_{\text 3}=(\text{I, II, V2})$, 
$\mathcal{C}_{\text 4}=(\text{I, II, III, V2})$,
$\mathcal{C}_{\text 6}=\text{(I, II, III, aVR, aVL, aVF)}$,
and $\mathcal{C}_{\text{12}}$ is the standard 12-leads~\cite{2021ChallengeCinC}.

{\sl EfficientNet} first generates a sequence of hidden representations $\bm{h}_{c, 1:T'} = (h_{c, t} \in {{\mathbb{R}}^{|H|}} | t=1 \cdots T', T' \leq T)$ by taking the ECG signal $X_c$, where $h_{c, t}$ is the $|H|$-dimensional hidden vector at frame $t$. 
These hidden representations are then passed to a global max-pooling layer to obtain a fixed-length representation $h_c$. We represent these neural network modules as follows:
\begin{align}
h_c = \text{pool}(f_{\text{effnet}}(X_c)),
\end{align}
where $\text{pool}(\cdot)$ and $f_{\text{effnet}}(\cdot)$ are the global max-pooling function and the {\sl EfficientNet} module.

In multi-class classification, the posterior probability of diagnoses is calculated using a softmax layer with additional fully connected layers (MLP):
\begin{align}
\hat{Y} = P(Y|X_c) = \text{softmax}(\text{MLP}(h_c)). \label{eq:ml_prediction}
\end{align}
We define $f_{\text{cls}} = \text{MLP}(\text{pool}(f_{\text{effnet}}(\text{X}_c)))$ for simplicity.
As our task is multi-label classification, we replace the softmax function with the sigmoid function in the next section.

\subsection{Division of Training Data}
Empirically, DNNs first learn to predict clean samples (expected to be high annotation quality), and later memorize noisy ones (expected to be poor annotation quality)~\cite{arpit2017closer}.
By exploiting this observation, {\sl DivideMix}~\cite{li2020dividemix} splits the training data into a set of clean samples and the one of noisy samples using two-component Gaussian Mixture Models (GMM).
In the training stage, two models, $f_{\text{cls}; \theta_m} (m=1, 2)$, are trained in parallel.
In other words, the training data that are split by $f_{\text{cls}; \theta_1}$ are used for the training of $f_{\text{cls}; \theta_2}$ in the next epoch, and vice versa.

In {\sl MixMatch}~\cite{berthelot2019mixmatch}, the noisy samples are used as unlabeled data.
In the case of multi-class classification, two networks make predictions using the softmax function, and the posterior probabilities are averaged to create a new label:
\begin{eqnarray}
{u}_{c, \theta_m}^{\text{nl}} &=& \text{softmax}(f_{\text{cls}; \theta_m}(\text{X}_c^{\text{nl}})), m=1, 2 \nonumber \\
{u}_c^{\text{nl}} &=& \text{Sharpen}(({u}_{c, \theta_1}^{\text{nl}} + {u}_{c, \theta_2}^{\text{nl}})/2.0), \label{eq:divmix_noisy}
\end{eqnarray}
where ${\text{X}_c^{\text{nl}}}$ is the ECG data estimated as a noisy label ($\text{nl}$) and $\text{Sharpen}$ is a function introduced in~\cite{li2020dividemix}.

\section{Proposed Method}
\vspace{-1mm}
\subsection{Multi-Label Label Refinement}
\vspace{-1mm}
In this section, we describe the modification of {\sl DivideMix} for the multi-label classification.
First, the $\text{softmax}$ function in Eq.~(\ref{eq:divmix_noisy}) is replaced with the $\text{sigmoid}$, and it is interpolated with the ground-truth label without the {\it sharpening} operation:
\begin{eqnarray}
{u}_{c, \theta_m}^{\text{nl}} &=& \text{sigmoid}(f_{\text{cls}; \theta_m}(X_c^{\text{nl}})), \nonumber \\
{u}_{c}^{\text{nl}} &=& \lambda_n ({u}_{c, 1}^{\text{nl}} + {u}_{c, 2}^{\text{nl}}) / 2.0 + (1-\lambda_n) Y,
\end{eqnarray}
where $\lambda_n$ is the interpolation coefficient, and $Y$ is the ground-truth label.
In the experiment, $\lambda_u$ was set to 0.5.
The label of clean sample $u_c^{\text{cl}}$ is updated as:
\begin{eqnarray}
u_{c, \theta_m}^{\text{cl}} = \lambda_{\text{gmm}} Y + (1-\lambda_{\text{gmm}}) \cdot \text{sigmoid}(f_{\text{cls}; \theta_{m}}(X_c^{\text{cl}})),
\end{eqnarray}
where $\lambda_{\text{gmm}}$ is the probability which is estimated as clean by GMM.
In contrast to the case of clean samples, the pseudo-label for the $m$-th network is estimated using the $m-$th network to reduce the training time.

Second, the sample-wise loss $l$ is updated to a binary cross-entropy loss $l_n$ and averaged over all labels:
\begin{eqnarray}
    l &=& \text{mean}(\left\{ l_1, \cdots, l_n \cdots, l_N\right\}), \\
    l_n &=& - \left\{Y_n \log(\hat{Y_n}) + (1-Y_n) \log(1-\hat{Y_n}), \right\}
\end{eqnarray}
where $Y$ and $\hat{Y}$ are the reference and estimated labels, respectively,  and $n$ is the label index.
Because the reduction of certain leads hinders the detection of one part of diagnostic characteristics but not all diagnoses, it is our future work to model label-dependent losses.

\subsection{Non-Sequential Manifold MixUp}
\vspace{-1mm}

Related works on image classification tasks apply {\sl MixMatch} to the input data domain.
However, it is not clear whether the interpolation of time-series data of different lengths affects the model training. 
Therefore, we generate the fixed-length hidden vector by using the max-pooling function and apply {\sl manifold-MixUp}~\cite{verma2019manifold}.

Let $\text{X}_{c}^{\text{cl}}$ and $Y^{\text{cl}}$ denote a pair of clean ECG  sample and reference label, $\text{X}_{c}^{\text{nl}}$ and $Y^{\text{nl}}$ denote a pair of noisy ECG sample and reference label.
The interpolation of the hidden vectors is then represented as follows:
\begin{eqnarray}
h_{c, \theta_m}^{\text{cl}} &=& \text{pool}(f_{\text{effnet}; \theta_m}(\text{X}_{c}^{\text{cl}})), \nonumber \\
h_{c, \theta_m}^{\text{nl}} &=& \text{pool}(f_{\text{effnet}; \theta_m}(\text{X}_{c}^{\text{nl}})), \nonumber \\
h_{c, \theta_m}^{\text{mix}} &=& \lambda_{\text{mix}} h_{c, \theta_m}^{\text{cl}} + (1-\lambda_{\text{mix}}) h_{c, \theta_m}^{\text{nl}},  \quad
\end{eqnarray}
where $\lambda_{\text{mix}}$ is the coefficient used in {\sl MixUp} sampled from a beta distribution. 
The interpolation of the label is:
\begin{eqnarray}
u_{c, \theta_m}^{\text{mix}} = \lambda_{\text{mix}} u_{c, \theta_m}^{\text{cl}} + (1-\lambda_{\text{mix}}) u_{c}^{\text{nl}},
\end{eqnarray}
and the objective function is defined as:
\begin{eqnarray}
{\mathcal{L}} = {L_x} + {L_u} &=& {\text{BCE}}(\text{MLP}({\mathcal{X}}(h_{c, \theta_m}^{\text{mix}})), {\mathcal{X}}({u}_{c, \theta_m}^{\text{mix}})) \nonumber \\
&+& {\text{L}}_{2}(\text{MLP}({\mathcal{U}}(h_{c, \theta_m}^{\text{mix}})), {\mathcal{U}}({u}_{c, \theta_m}^{\text{mix}})) ,\label{eq:bce_loss}
\end{eqnarray}
where $\text{BCE}$ and $\text{L}_2$ are the binary cross-entropy and mean squared loss functions with the sigmoid function, and $\mathcal{X}$ and $\mathcal{U}$ are
dummy functions which divide the samples into clean/noisy samples.

\subsection{Model Ensemble}
\vspace{-1mm}
The model ensemble is a technique that combines predictions calculated by multiple classifiers for variance reduction (discussed in a context of a bias-variance trade-off).
Under the proposed framework, the two models were trained in parallel. Therefore, these two models can be used for model ensemble.

Stochastic weight averaging (SWA)~\cite{izmailov2018averaging} creates a new model by averaging the model weights sampled at different stages of training.
In the experiment, we applied SWA to the two models to generate two averaged models.
The result of final prediction is an arithmetic mean of the posterior probabilities calculated by the four models.

\subsection{Model Architecture}
\vspace{-1mm}
\label{lab:model_arch}

\begin{table}[tb]
    \centering
    \caption{{\sl EfficientNet} model architecture. Each line describes a sequence of 1D convolution layer or Fused-MBConv (mobile inverted residual bottleneck convolution) modules consists of k-size kernels. The first convolutional layer of each stage has stride shown in the 3rd column and the followings use 1. \#Channels is the number of output channels of each stage.}
    \small
    \label{tab:effnet_arch}
    \begin{tabular}{c|cccc} \hline
        Stage & Operator & Stride & \#Channels & \#Layers \\ \hline
        0 & Conv, k: 7 & 2 & 32 & 1 \\
        1 & Fused-MBConv2, k: 5 & 2 & 32 & 2 \\
        2 & Fused-MBConv1, k: 5 & 2 & 64 & 1 \\
        3 & Fused-MBConv2, k: 7 & 2 & 128 & 2 \\
        4 & Fused-MBConv1, k: 7 & 2 & 128 & 1 \\
        5 & Fused-MBConv2, k: 7 & 2 & 256 & 2 \\
        6 & Fused-MBConv2, k: 7 & 2 & 256 & 2 \\ 
        7 & Conv, k: 1 & 1 & 512 & 1 \\ \hline
    \end{tabular}
    \vspace{-5mm}
\end{table}

Table~\ref{tab:effnet_arch} shows the model architecture based on EfficientNet~\cite{tan2021efficientnetv2}.
Fused-MBConv is a sequence of 1) 1D convolution layer, 2) squeeze-and-excitation module, and 3) point-wise convolution layer. 
1) The input tensor (W, C) is expanded to (W', 2C) at the first convolution layer followed by batch normalization (BN) and the Mish function~\cite{misra2019mish} where W, C are the width and channel sizes.
2) In the squeeze-and-excitation module, channel statistics are summarized by a pooling function, and its dimension is reduced to C/4.
This embedded feature is expanded to C followed by a sigmoid function for channel-wise attention.
3) Lastly, point-wise convolution and BN are used to update the output channel size.

Natarajan, et al.,~\cite{natarajan2020wide} proposed {\sl wide}-and-{\sl deep} Transformer neural networks.
This approach uses a Transformer network to compute a fixed-length representation. It is fused with hand-crafted ECG features on top of the Transformer network to incorporate expert knowledge.
Likewise, we used age, gender, and RR-interval-related features extracted from lead II as the {\sl wide} features. 
These {\sl wide} features are concatenated before the point-wise convolution to condition the all Fused-MBConv blocks.

\section{Experimental Setup and Results}
\vspace{-1mm}

\subsection{Feature Extraction}

We used the CPSC database~\cite{CPSC}, INCART database~\cite{INCART}, PTB database~\cite{PTB}, PTB-XL database~\cite{PTB-XL}, Chapman-Shaoxing Database~\cite{Chapman-Shaoxing}, Ningbo Database~\cite{Ningbo}, and other databases~\cite{2020ChallengePMEA, 2021ChallengeCinC}.

All ECG signals were resampled to 500 Hz and normalized to a range of $[-1, 1]$ by min-max normalization for each lead.
We extracted 15 seconds of ECG data from a random starting point and applied zero-padding when the duration was shorter than 15 seconds. 
When the duration (before zero-padding) $\tau$ was longer than 10 seconds, we decreased its duration randomly by sampling from the uniform distribution U(10, $\tau$) to make the network learn duration-independent prediction.

We used stratified 10-fold cross-validation and averaged over 10 challenge metric scores for each reduced (2, 3, 4, 6, and 12) leads setup~\cite{2021ChallengeCinC} to test the effectiveness of the proposed method. 
No additional processing was added to the different lead combinations.
The Welch t-test was used for the statistical test.

\vspace{-1mm}
\subsection{Optimization}
\vspace{-1mm}

\begin{itemize}
    \item Baseline Model: We used the model described in Section~\ref{lab:model_arch}.
    The number of output units was set to 24 which corresponds to diagnoses scored by the Physionet 2021 Challenge.
    We used the Adam algorithm~\cite{kingma2014adam} 
    and minimized the binary cross-entropy loss.
    The model was trained for 40 epochs with a batch size of 240.
    As the {\sl wide} feature, we extracted age, gender, and RR-interval-related features computed by  biosppy~\cite{carreiras2015biosppy} and hrv~\cite{bartels2020hrv}.
    It is passed to 4-layer fully connected layers with BN and a Mish function~\cite{misra2019mish} followed by the Fused-MBConv module.
    2-layer fully connected layers were used as the MLP introduced in Eq.~(\ref{eq:bce_loss}). The predicted posterior probabilities were converted to positive or negative based on a fixed threshold of 0.3.
    
    \item Proposed Model: 
    The model was trained for 40 epochs with a batch size of 160.
    The first two epochs were trained as the baseline model, and the other epochs were trained under the proposed framework. 
    SWA was applied for the last 13 epochs.
    The number of expectation-maximization algorithm iterations used for GMM training was set to 10. All the models were trained from scratch.
\end{itemize}

\subsection{Results}
\vspace{-1mm}
\begin{table}[tbp]
    \centering
    \small
    \caption{Challenge scores for our final selected entry (team ami\_kagoshima) using 10-fold cross validation on the public training set, repeated scoring on the hidden validation set, and one-time scoring on the hidden test set as well as the ranking on the hidden test set.}
    \begin{tabular}{r|r|r|r|r}
        Leads & Training        & Validation & Test & Ranking \\\hline
        12    & $0.701 \pm 0.006$ &       0.623 &  0.49 &     9 \\
         6    & $0.686 \pm 0.003$ &       0.593 &  0.47 &     10 \\
         4    & $0.693 \pm 0.006$ &       0.606 &  0.48 &     10 \\
         3    & $0.693 \pm 0.005$ &       0.612 &  0.47 &     11 \\
         2    & $0.685 \pm 0.006$ &       0.601 &  0.47 &     10 \\\hline
    \end{tabular}
    \label{tab:scores}
    \vspace{-4mm}
\end{table}

The averaged challenge-scores of the baseline method were
0.682,  
0.667,  
0.676,  
0.673, and  
0.664  
on the 12-, 6-, 4-, 3- and 2-leads ECG data, respectively.
Table~\ref{tab:scores} shows the challenge scores of the proposed method.
Our results on 10-fold cross-validation were
$0.701 (2.8\%)^{(***)}$, 
$0.686 (2.8\%)^{(***)}$, 
$0.693 (2.5\%)^{(**)}$, 
$0.693 (3.0\%)^{(***)}$, and 
$0.685 (3.2\%)^{(***)}$ 
on the 12-, 6-, 4-, 3- and 2-leads ECG data, respectively\footnote{***: p < 0.001, **: p < 0.01}.
The values given in the parentheses represent relative improvements.
The proposed method obtained the score of 0.49, 0.47, 0.48, 0.47, and 0.47 for each lead combination on the hidden test set.

\section{Discussion and Conclusion}
In this paper, we have proposed reduced-lead ECG classifiers based on {\sl DivideMix} and SWA.
As the reduction of certain ECG leads hinders the cardiac electrical signal, it is expected to degrade the classification performance when the available leads are limited.
We can see that the challenge scores of the baseline and proposed models decreased linearly except for the 6-leads setup.
The proposed method have obtained relatively large improvements on 2- and 3-leads setups.
It is considered that the proposed method alleviated performance degradation owing to the poor annotation quality.
Future work is detailed diagnoses-level investigations of the performance changes caused by the reduction of available lead combinations.

\bibliographystyle{cinc}
\bibliography{references}

\vspace{-5mm}
\begin{correspondence}
Hiroshi Seki (hseki@ami.inc) \\
302, 2-13 Higashi-Sengoku, Kagoshima, Japan 
\end{correspondence}

\vspace{-5mm}
\end{document}